# Polyharmonic Cascade

Bakhvalov Y. N., Ph.D., Independent Researcher, Cherepovets, Russia

bahvalovj@gmail.com, ORCID: 0009-0002-5039-2367

**Abstract.**

This paper presents a deep machine learning architecture, the "polyharmonic cascade" – a sequence of packages of polyharmonic splines [2], where each layer is rigorously derived from the theory of random functions and the principles of indifference [1]. This makes it possible to approximate nonlinear functions of arbitrary complexity while preserving global smoothness and a probabilistic interpretation.

For the polyharmonic cascade, a training method alternative to gradient descent is proposed: instead of directly optimizing the coefficients, one solves a single global linear system on each batch with respect to the function values at fixed "constellations" of nodes. This yields synchronized updates of all layers, preserves the probabilistic interpretation of individual layers and theoretical consistency with the original model from the first paper in the series [1], and scales well: all computations reduce to 2D matrix operations efficiently executed on a GPU. Fast learning without overfitting on MNIST is demonstrated.

This paper is devoted to scaling up the solution obtained in [1] for approximating multidimensional nonlinear functions of unbounded complexity. In [1] it was shown that a machine learning regression problem can be solved (and admits a closed-form solution) within the framework of the theory of random functions [12] and the principles of indifference. The formulas derived in [1] can, with certain nuances, be interpreted as a special case of a polyharmonic spline, namely a thin-plate spline [10], [5].

In [2] it was shown that if, in the solution from [1], one specifies a fixed number of key points (called a "constellation"), then within this constellation one can combine a large number of functions (polyharmonic splines) into a single package (the notion of a package was introduced in [2]), which can be evaluated and differentiated efficiently.

It was also shown in [2] that the solution from [1] has limitations of applicability, both due to the rapid growth of computational complexity for large training sets and due to obtaining suboptimal results in certain cases analyzed in [2]. The same limitations apply to packages of such functions.

To remove these limitations, [2] proposed (and provided justification for) connecting a sequence of packages of polyharmonic splines into a multilayer computational structure. Expressions were given both for the forward computation of such a structure (sequentially from the first layer to the last) and for its differentiation (moving in reverse order from the last layer to the first using the chain rule).

All expressions were obtained in the form of matrix operations, which makes them relatively easy to implement and efficient to compute on a GPU.

We will refer to such a computational structure – a sequential, cascaded connection of packages of polyharmonic functions – as a polyharmonic cascade.

There is an obvious analogy between the polyharmonic cascade and artificial multilayer neural networks, both in terms of their structure (a sequence of layers) and their purpose. However, if one treats this computational structure (a sequence of polyharmonic function packages) as just another variety of multilayer neural network, and calls it, for example, a "polyharmonic neural network", such a name may create misleading associations regarding the nature and operating principles of the algorithm.

There are substantial differences between the polyharmonic cascade and standard multilayer neural network algorithms. This concerns its mathematical representation and justification. The basic mathematical expressions underlying the polyharmonic cascade can be derived from the theory of random functions ([12] and [1]) and the principles of indifference. Mathematically, a polyharmonic spline function differs significantly from the (commonly adopted) mathematical model of a neuron, and a polyharmonic package differs from a layer of neurons.

Although compositions of Gaussian processes and kernel functions have been studied in the literature on Deep Gaussian Processes (Damianou & Lawrence, 2013 [7]) and Deep Kernel Learning (Wilson et al., 2016 [16]), the polyharmonic cascade differs from these approaches in that each layer retains a strict probabilistic justification via symmetry and indifference principles. Polyharmonic splines form a canonical class of radial basis functions with optimal smoothness and invariance properties (Buhmann 2003 [6]; Duchon 1977 [9]). Unlike arbitrary RBFs (e.g., Gaussian), they do not impose an a priori preference for low-frequency components, which is crucial in the absence of prior information. Artificially grouping arbitrary radial basis functions (other than the polyharmonic spline) into a package over a given constellation can lead to oscillations and make the algorithm fail (this depends on the spectral decomposition of the radial basis function, as shown in [1]).

Perhaps the most important distinguishing feature of the polyharmonic cascade is its training method, which is fundamentally different from that of neural networks yet remains effective. The point is that one cannot straightforwardly train it by gradient descent directly over the equation coefficients (without additional tricks). It is easy to define a computation graph for the polyharmonic cascade (including in any modern machine learning framework) and to compute error gradients with respect to all its coefficients. However, applying any optimization procedure based on following the gradient direction to these coefficients themselves is practically ineffective. The parameter space formed by the tunable coefficients of the polyharmonic cascade is extremely ill-suited for gradient-descent-based optimization.

Nevertheless, an efficient training procedure for a polyharmonic cascade (with many layers) does exist and will be developed in this paper.

Let us represent the polyharmonic cascade schematically. Suppose we have a cascade of q polyharmonic packages connected in sequence (Figure 1). The diagram shows the first and last packages in the cascade, as well as some intermediate package with index $\tau$.

At the input of the first package we receive a batch of data to be processed, in the form of a matrix $X_0$ of size $(r \times n_0)$, where each row is an independent input vector of dimension $n_0$ (which is the number of inputs of the computational system and the dimensionality of the function ultimately computed by the polyharmonic cascade). The number of vectors in this batch (which we treat as a training batch) is $r$.

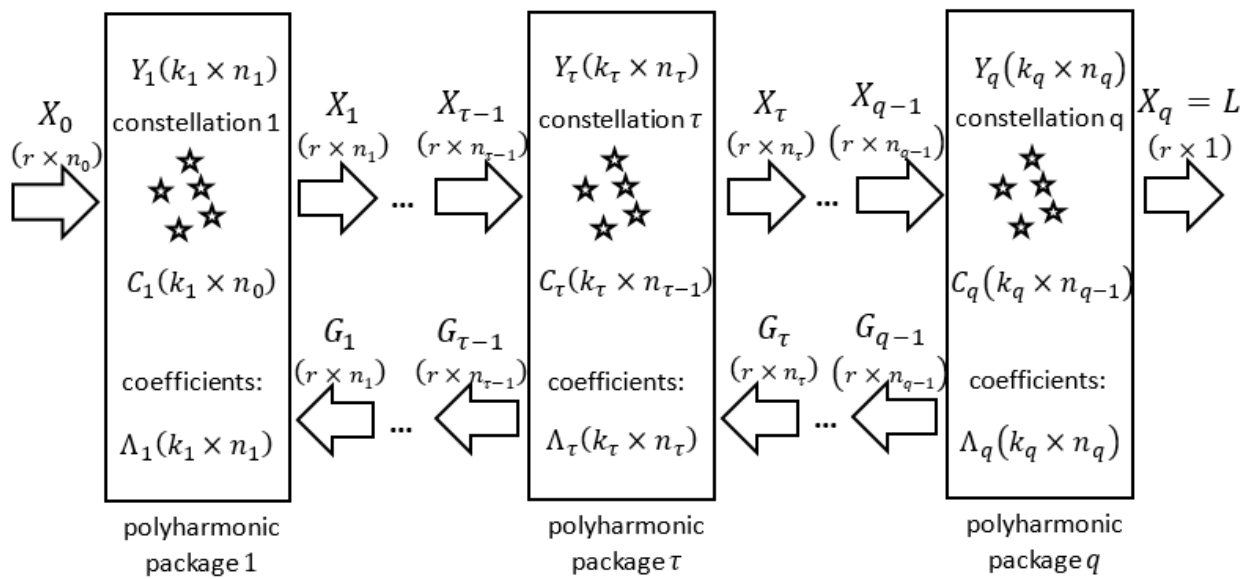

Figure 1. Polyharmonic cascade

At the output of the first polyharmonic package we obtain a matrix $X_1$ of size $(r \times n_1)$, where $n_1$ is the number of outputs of the first package (i.e., the number of functions it computes). This matrix is then fed into the second polyharmonic package, and so on. For the package with index $\tau$ the corresponding input and output matrices are $X_{\tau-1}$ and $X_\tau$, respectively.

At the output of the cascade, from the last package, we obtain a matrix $X_q$ of size $(r \times 1)$, which is in fact a vector of length $r$ that we denote by $L = (l_1, l_2, \ldots, l_r)$, where each component $l_i$ is the final result produced by the entire polyharmonic cascade for the i-th row of the original matrix $X_0$. That is, we consider a cascade that has a single output at the last layer and, as a whole, computes a single function. After describing the training algorithm for this case, we will discuss how it can be modified when there are multiple outputs.

The polyharmonic cascade also allows us to traverse it in the reverse direction and compute the derivative matrices $G_1, G_2, \ldots, G_{q-1}$ (this procedure was described in [2]), in a way very similar to backpropagation in neural networks ([3] and [13]). However, in this case we are not dealing with the gradient of a loss function, but with derivatives of the very function computed by the polyharmonic cascade. Secondly, these derivative matrices are computed not with respect to the equation coefficients (whose adjustment is the ultimate goal), but with respect to the outputs of the packages, i.e., with respect to the values in the matrices $X_1, X_2, \ldots, X_{q-1}$.

Thus, the elements of, for example, the matrix $G_\tau$ can be written as $g_{it}^{(\tau)} = \frac{dl_i}{dx_{it}^{(\tau)}}$, where $\tau = \overline{1, q-1}$, $i = \overline{1, r}$, $t = \overline{1, n_\tau}$.

Accordingly, the sizes of the matrices $G_1, G_2, \ldots, G_{q-1}$ coincide exactly with the sizes of $X_1, X_2, \ldots, X_{q-1}$.

In the figure, the matrices $G_0$ and $G_q$ corresponding to $X_0$ and $X_q$ are not shown. The matrix $G_0$ can in principle be computed, but it is not used in the training algorithm described below. The matrix $G_q$, on the other hand, will be used; clearly it has size $(r \times 1)$ and is simply a vector of length $r$ filled with ones (since in this case $dx_{i1}^{(q)} = dl_i$).

Each polyharmonic package in the diagram is described by three matrices. For the package with index $\tau$ these are $Y_\tau, C_\tau$, and $\Lambda_\tau$. The matrix $C_\tau$ of size $(k_\tau \times n_{\tau-1})$ specifies the constellation points in the package, the matrix $Y_\tau$ of size $(k_\tau \times n_\tau)$ contains the function values at these points, and the matrix $\Lambda_\tau$ of size $(k_\tau \times n_\tau)$ holds the coefficients that are directly used in computation (although they are derived from $C_\tau$ and $Y_\tau$).

We now recall all formulas obtained in [1] and [2] that are needed to perform the computations shown in the diagram and those associated with them.

First, we describe how a polyharmonic package with index $\tau$ computes the matrix $X_\tau$ from the input matrix $X_{\tau-1}$.

As a first step, we compute the matrix of all pairwise squared distances between the row vectors in $X_{\tau-1}$ and the constellation points (denote it by $M_\tau$):

$$M_\tau = N_{\tau x} J_{1,k_\tau} + J_{r,1} N_{\tau c}^T - 2X_{\tau-1} C_\tau^T, \tag{1}$$

where

$$N_{\tau x} = (X_{\tau-1} \circ X_{\tau-1}) J_{n_{\tau-1},1}, \tag{2}$$

$$N_{\tau c} = (C_\tau \circ C_\tau) J_{n_{\tau-1},1}, \tag{3}$$

"∘" denotes the Hadamard (elementwise) product,

$J_{1,k_\tau}$ is a $1 \times k_\tau$ row vector of ones,

$J_{r,1}$ is an $r \times 1$ column vector of ones,

$J_{n_{\tau-1},1}$ is an $n_{\tau-1} \times 1$ column vector of ones,

$C_\tau$ is the constellation matrix of the $\tau$-th package.

In fact, (1) is just the law of cosines written in matrix form.

Next, from $M_\tau$ we obtain a matrix $K_\tau$ of the same size ($r \times k_\tau$) by transforming each of its elements:

$$k_{ip}^\tau = m_{ip}^\tau \left( \ln(m_{ip}^\tau) - b \right) + c, \tag{4}$$

where

b and c are coefficients whose values were estimated in [1],

$k_{ip}^\tau$ is an element of $K_\tau$ (not to be confused with $k_\tau$),

$m_{ip}^\tau$ is the corresponding element of $M_\tau$,

$i = \overline{1,r}, p = \overline{1,k_\tau}$.

Compared to its use in [1] and [2], expression (4) is slightly simplified: the factor 1/2 in front of the expression and the factor 2 in front of b have been removed. This does not change anything essential, since b and c themselves can be chosen accordingly. In numerical experiments with (4), the values b = 10 and c = 1000 worked well.

Next, we obtain $X_\tau$ as the matrix product of $K_\tau$ and $\Lambda_\tau$:

$$X_\tau = K_\tau \Lambda_\tau \tag{5}$$

We now turn to the procedure for propagating derivatives: how to compute $G_{\tau-1}$ given $G_\tau$.

First, we compute the matrix $\Theta_\tau$. It is obtained by an elementwise transformation of the matrix $M_\tau$ from (1), using

$$\theta_{ip}^{\tau} = \ln(m_{ip}^{\tau}) - b + 1, \qquad (6)$$

where

$b$ is the same coefficient as in (4),

$\theta_{ip}^{\tau}$ is an element of $\Theta_{\tau}$,

$m_{ip}^{\tau}$ is an element of $M_{\tau}$,

$i = \overline{1, r}, p = \overline{1, k_{\tau}}$.

Then we compute the matrix $\Psi_{\tau}$:

$$\Psi_{\tau} = \Theta_{\tau} \circ (G_{\tau} \Lambda_{\tau}^{T}) \qquad (7)$$

where "∘" denotes the Hadamard (elementwise) product.

The matrices $\Theta_{\tau}$ and $\Psi_{\tau}$ both have size $(r \times k_{\tau})$.

We can now express $G_{\tau-1}$ as

$$G_{\tau-1} = X_{\tau-1} \circ \left(\Psi_{\tau} J_{k_{\tau},1} J_{1,n_{\tau-1}}\right) - \Psi_{\tau} C_{\tau}, \qquad (8)$$

where

$J_{k_{\tau},1}$ is a $k_{\tau} \times 1$ column vector of ones,

$J_{1,n_{\tau-1}}$ is a $1 \times n_{\tau-1}$ row vector of ones.

Thus, in the polyharmonic cascade, expressions (1)-(5) define the forward computations, while (6)-(8) define the backward derivative propagation.

Before moving on to the training algorithm, let us show that the forward computations can be written in an alternative form. In (5) we use the matrix $\Lambda_{\tau}$, which in turn is determined by $C_{\tau}$ and $Y_{\tau}$. We now show how $\Lambda_{\tau}$ is computed.

First, we compute the matrix of all pairwise squared distances between the constellation points themselves; denote it by $M_{\tau}^{(C)}$:

$$M_{\tau}^{(C)} = N_{\tau c} J_{1,k_{\tau}} + J_{k_{\tau},1} N_{\tau c}^{T} - 2 C_{\tau} C_{\tau}^{T}, \qquad (9)$$

where

$N_{\tau c} = (C_{\tau} \circ C_{\tau}) J_{n_{\tau-1},1}$ (the same expression as in (3)),

$J_{1,k_{\tau}}$ is a $1 \times k_{\tau}$ row vector of ones,

$J_{k_{\tau},1}$ is a $k_{\tau} \times 1$ column vector of ones.

We then define the matrix $U_\tau$ as

$$U_\tau = \left(K_\tau^{(C)} + \sigma_\tau^2 E\right)^{-1}, \tag{10}$$

where

$K_\tau^{(C)}$ is computed from $M_\tau^{(C)}$ using (4),

$\sigma_\tau^2$ is the variance of the random variables (its meaning is discussed in more detail in [1]),

E is the identity matrix.

The value $\sigma_\tau^2$ plays the role of a "regularization coefficient", turning interpolation into approximation and back (depending on its value) when computing the functions from the constellation points within a package. Therefore, its value may depend on how and according to what principle the points in the constellation were chosen. If duplicate points in the constellation are excluded and the points are reasonably well distributed, $\sigma_\tau^2$ can be taken close to zero or even set to zero. In this case, within the package, relative to the values in $Y_\tau$, we are effectively performing interpolation, which, however, does not preclude approximation with respect to the training set by the polyharmonic cascade as a whole.

The matrix $\Lambda_\tau$ is computed as the product of $U_\tau$ and $Y_\tau$:

$$\Lambda_\tau = U_\tau Y_\tau \tag{11}$$

Using (5) and (11) we obtain

$$X_\tau = K_\tau \Lambda_\tau = K_\tau U_\tau Y_\tau \tag{12}$$

If we additionally introduce the matrix

$$H_\tau = K_\tau U_\tau \tag{13}$$

then (12) can be rewritten as

$$X_\tau = H_\tau Y_\tau \tag{14}$$

Within the operation and training algorithm for the polyharmonic cascade, we will assume that the locations of the constellation points in each package (the matrices $C_\tau$) are specified at initialization in some way and remain fixed thereafter (we treat their values as constants).

Although, in principle, the entries of the matrices $C_1, C_2, .., C_q$ could also be adjusted during training, numerical experiments showed that fixing them does not prevent the polyharmonic cascade from learning effectively. The initialization algorithm for $C_\tau$ was the subject of a separate study, and its results will be presented in a separate paper. Fixing the matrices $C_\tau$ has an additional advantage: in this case the matrix $U_\tau$ from (10) for each polyharmonic package also remains constant, and expressions (9) and (10) need to be evaluated only once. Nevertheless, algorithms in which the values in $C_1, C_2, .., C_q$ are also updated during training require further investigation.

Thus, expressions (12)-(14), which can be used in place of (5), show that the polyharmonic cascade can be represented in such a way (slightly more cumbersome, involving the matrices $U_\tau$) that the equation coefficients (the matrices $\Lambda_\tau$) are removed from the explicit computation and their role is taken over by the function values (the matrices $Y_\tau$) at constellation points.

Under the representation (12)-(14) and with $C_\tau$ fixed, training the polyharmonic cascade reduces to finding appropriate values of $Y_1, Y_2, .., Y_q$, i.e., to correctly specifying the values of the functions in each polyharmonic package at its constellation points.

Now, if we denote the entire cascade as the evaluator of some function $F$, then the computation of a particular output value $l_i$ in the vector $L$ can be written as

$$l_i = F\left(x_i^{(0)}, Y_1, Y_2, .., Y_q\right), \tag{15}$$

where

$x_i^{(0)}$ is the i-th row of the matrix $X_0$.

In (15) we thus treat the matrices $Y_1, Y_2, .., Y_q$ as additional input parameters, as if they were fed into the cascade, and regard the cascade itself as a fixed function.

Suppose we wish to change $Y_1, Y_2, .., Y_q$ in such a way that, for the same input $x_i^{(0)}$, the output becomes a new desired value $l_i^*$. If we can find changes $\Delta Y_1, \Delta Y_2, .., \Delta Y_q$ such that this condition is satisfied, we can write

$$l_i^* = F\left(x_i^{(0)}, Y_1 + \Delta Y_1, Y_2 + \Delta Y_2, .., Y_q + \Delta Y_q\right) \tag{16}$$

Since $F$ is differentiable with respect to every element of each matrix $Y_1, Y_2, .., Y_q$, expression (16) can be expanded as

$$F\left(x_i^{(0)}, Y_1 + \Delta Y_1, Y_2 + \Delta Y_2, .., Y_q + \Delta Y_q\right) =$$
$$= F\left(x_i^{(0)}, Y_1, Y_2, .., Y_q\right) + \sum_{\tau=1}^{q}\sum_{p=1}^{k_\tau}\sum_{t=1}^{n_\tau}\left(\frac{dl_i}{dy_{pt}^{(\tau)}}\Delta y_{pt}^{(\tau)}\right) + e_i, \tag{17}$$

where

$y_{pt}^{(\tau)}$ are the elements of $Y_\tau$,

$\Delta y_{pt}^{(\tau)}$ are the elements of $\Delta Y_\tau$,

$e_i$ is the approximation error.

Thus we have replaced the function in (16) by the sum of the original function value from (15) and the value of its tangent hyperplane at the point $x_i^{(0)}$. If $F$ is continuous and differentiable (as is the function implemented by the polyharmonic cascade), then for infinitesimally small $\Delta y_{pt}^{(\tau)}$ this representation is exact. For larger values of $\Delta y_{pt}^{(\tau)}$, a discrepancy appears, which is captured by the error term $e_i$ in (17).

How can we compute $\dfrac{dl_i}{dy_{pt}^{(\tau)}}$ from (17)?

Expression (14), written as a matrix product, can be equivalently expressed elementwise as

$$x_{it}^{(\tau)} = \sum_{p=1}^{k_\tau} h_{ip}^{(\tau)} y_{pt}^{(\tau)}, \qquad (18)$$

where

$x_{it}^{(\tau)}$ are the elements of the matrix $X_\tau$,

$h_{ip}^{(\tau)}$ are the elements of the matrix $H_\tau$,

$y_{pt}^{(\tau)}$ are the elements of the matrix $Y_\tau$.

From (18) we obtain $\dfrac{dx_{it}^{(\tau)}}{dy_{pt}^{(\tau)}} = h_{ip}^{(\tau)}$.

As already noted, we have $\dfrac{dl_i}{dx_{it}^{(\tau)}} = g_{it}^{(\tau)}$, where $g_{it}^{(\tau)}$ is an element of the matrix $G_\tau$. Hence,

$$\frac{dl_i}{dy_{pt}^{(\tau)}} = \frac{dl_i}{dx_{it}^{(\tau)}} \frac{dx_{it}^{(\tau)}}{dy_{pt}^{(\tau)}} = g_{it}^{(\tau)} h_{ip}^{(\tau)} \qquad (19)$$

Using (16), (17), and (19), and writing these equations for the cascade processing each row $x_1^{(0)}, x_2^{(0)}, \ldots, x_r^{(0)}$ of the matrix $X_0$, we obtain a system of $r$ equations:

$$\begin{cases} \sum_{\tau=1}^{q}\sum_{p=1}^{k_\tau}\sum_{t=1}^{n_\tau} \left(g_{1t}^{(\tau)} h_{1p}^{(\tau)} \Delta y_{pt}^{(\tau)}\right) + e_1 = l_1^* - F\left(x_1^{(0)}, Y_1, Y_2, \ldots, Y_q\right) \\ \qquad\qquad\qquad \ldots \\ \sum_{\tau=1}^{q}\sum_{p=1}^{k_\tau}\sum_{t=1}^{n_\tau} \left(g_{rt}^{(\tau)} h_{rp}^{(\tau)} \Delta y_{pt}^{(\tau)}\right) + e_r = l_r^* - F\left(x_r^{(0)}, Y_1, Y_2, \ldots, Y_q\right) \end{cases} \quad (20)$$

Or, equivalently,

$$\begin{cases} \sum_{\tau=1}^{q}\sum_{p=1}^{k_\tau}\sum_{t=1}^{n_\tau} \left(g_{1t}^{(\tau)} h_{1p}^{(\tau)} \Delta y_{pt}^{(\tau)}\right) + e_1 = l_1^* - l_1 = \Delta l_1 \\ \qquad\qquad\qquad \ldots \\ \sum_{\tau=1}^{q}\sum_{p=1}^{k_\tau}\sum_{t=1}^{n_\tau} \left(g_{rt}^{(\tau)} h_{rp}^{(\tau)} \Delta y_{pt}^{(\tau)}\right) + e_r = l_r^* - l_r = \Delta l_r \end{cases}, \quad (21)$$

where
$\Delta l_i$ is the i-th element of $\Delta L = L^* - L$,
$L^*$ is the vector of desired outputs $l_i^*$ at the cascade output.

Having obtained (21), we can now formulate an optimization problem: find such minimal values of $\Delta y_{pt}^{(\tau)}$ (i.e., change the values at the constellation points, represented by the matrices $Y_1, Y_2, \ldots, Y_q$, as little as possible) that the errors $e_1, e_2, \ldots, e_r$ are minimized, subject to the constraints (21). In other words, we must balance, on one hand, minimizing the errors, and on the other hand keeping the magnitudes of $\Delta y_{pt}^{(\tau)}$ as small as possible.

Note that in this formulation the errors $e_1, e_2, \ldots, e_r$ have two components: they represent both the mismatch between the cascade outputs and the desired values $l_i^*$, and the approximation error introduced by replacing the model with its linearization in (17), whose magnitude is unknown. Therefore, trying to drive $e_1, e_2, \ldots, e_r$ to zero in a single iteration may actually harm the learning process.

These goals can be achieved by minimizing the objective function

$$\frac{1}{\alpha}\sum_{i=1}^{r} e_i^2 + \sum_{\tau=1}^{q}\sum_{p=1}^{k_\tau}\sum_{t=1}^{n_\tau} \left(\Delta y_{pt}^{(\tau)}\right)^2 \to \min \quad (22)$$

where the coefficient $\alpha$ controls the trade-off between minimizing the errors $e_i$ and minimizing the changes $\Delta y_{pt}^{(\tau)}$.

Taken together, (21) and (22) define a quadratic programming problem with equality constraints. This problem is solved using the method of Lagrange multipliers.

At first glance, examining (21) and (22) may raise doubts about the practicality of the proposed approach. Each equation in (21) contains as many coefficients as there are trainable parameters in the entire polyharmonic cascade, and the number of equations is $r$ – the number of training examples in the batch. If, for example, the batch contains one thousand examples, then simply storing all the coefficients of (21) in matrix form would require roughly a thousand times more memory than the number of trainable parameters in the cascade itself. One might also expect that solving such a system would demand substantial computational resources. However, as will be shown below, these concerns are unfounded.

Let us write the Lagrangian and denote it by $\hat{L}$, to distinguish it from the previously used symbol $L$ for the vector of cascade outputs. We denote the Lagrange multipliers by $\beta_i$ (instead of $\lambda_i$), and collect them into a vector $B$, so as not to confuse them with the matrix $\Lambda$ and its elements already in use.

The Lagrangian has the form

$$\hat{L}\left(e_1, e_2, \ldots, e_r, \Delta y_{11}^{(1)}, \ldots, \Delta y_{pt}^{(\tau)}, \ldots, \Delta y_{k_q n_q}^{(q)}, \beta_1, \beta_2, \ldots, \beta_r\right) =$$

$$= \frac{1}{\alpha} \sum_{i=1}^{r} e_i^2 + \sum_{\tau=1}^{q} \sum_{p=1}^{k_\tau} \sum_{t=1}^{n_\tau} \left(\Delta y_{pt}^{(\tau)}\right)^2 +$$

$$+ \sum_{i=1}^{r} \beta_i \left( \Delta l_i - e_i - \sum_{\tau=1}^{q} \sum_{p=1}^{k_\tau} \sum_{t=1}^{n_\tau} \left(g_{it}^{(\tau)} h_{ip}^{(\tau)} \Delta y_{pt}^{(\tau)}\right) \right) \quad (23)$$

If in (23) we set $\frac{d\hat{L}}{d\beta_i} = 0$, we recover the constraint system (21).

If we set $\frac{d\hat{L}}{de_i} = 0$, we obtain the relationship between $e_i$ and $\beta_i$ in the solution:

$$e_i = \frac{\alpha}{2} \beta_i \quad (24)$$

If we set $\frac{d\hat{L}}{d\Delta y_{pt}^{(\tau)}} = 0$, we obtain

$$\Delta y_{pt}^{(\tau)} = \frac{1}{2} \sum_{i=1}^{r} \beta_i g_{it}^{(\tau)} h_{ip}^{(\tau)} \quad (25)$$

Combining (21), (24), and (25), we arrive at the system

$$\begin{cases} \sum_{\tau=1}^{q}\sum_{p=1}^{k_\tau}\sum_{t=1}^{n_\tau}\left(g_{1t}^{(\tau)}h_{1p}^{(\tau)}\sum_{i=1}^{r}\beta_i g_{it}^{(\tau)}h_{ip}^{(\tau)}\right) + \alpha\beta_1 = 2\Delta l_1 \\ \quad\quad\quad\quad\quad\quad \ldots \\ \sum_{\tau=1}^{q}\sum_{p=1}^{k_\tau}\sum_{t=1}^{n_\tau}\left(g_{rt}^{(\tau)}h_{rp}^{(\tau)}\sum_{i=1}^{r}\beta_i g_{it}^{(\tau)}h_{ip}^{(\tau)}\right) + \alpha\beta_r = 2\Delta l_r \end{cases} \quad (26)$$

where $\Delta l_i$ are the components of $\Delta L$.

Thus, solving the optimization problem reduces to solving the system (26) to find $\beta_1, \beta_2, \ldots, \beta_r$. Once these are known, any $\Delta y_{pt}^{(\tau)}$ can be computed from (25), i.e., we can obtain the update matrices $\Delta Y_1, \Delta Y_2, \ldots, \Delta Y_q$ for the constellation values.

In (26) we can rearrange the order of summation and rewrite the system as

$$\begin{cases} \sum_{i=1}^{r}\beta_i \sum_{\tau=1}^{q}\left(\sum_{p=1}^{k_\tau}h_{1p}^{(\tau)}h_{ip}^{(\tau)}\right)\left(\sum_{t=1}^{n_\tau}g_{1t}^{(\tau)}g_{it}^{(\tau)}\right) + \alpha\beta_1 = 2\Delta l_1 \\ \quad\quad\quad\quad\quad\quad \ldots \\ \sum_{i=1}^{r}\beta_i \sum_{\tau=1}^{q}\left(\sum_{p=1}^{k_\tau}h_{rp}^{(\tau)}h_{ip}^{(\tau)}\right)\left(\sum_{t=1}^{n_\tau}g_{rt}^{(\tau)}g_{it}^{(\tau)}\right) + \alpha\beta_r = 2\Delta l_r \end{cases} \quad (27)$$

which can be compactly expressed in matrix form as

$$\left(\sum_{\tau=1}^{q}\left((H_\tau H_\tau^T)\circ(G_\tau G_\tau^T)\right) + \alpha E\right)B = 2\Delta L, \quad (28)$$

where
$E$ is the $r \times r$ identity matrix,
$B$ is the column vector $\beta_1, \beta_2, \ldots, \beta_r$,
and "∘" denotes the Hadamard (elementwise) product.

It is precisely the possibility of transforming (26) into (27) and then into (28) that allows us, in this problem, to drastically reduce both the memory required and the amount of computation, compared to solving a generic quadratic programming problem where the constraint coefficients are arbitrary.

Let us denote the summand in (28) for each layer by a separate matrix:

$$\Omega_\tau = (H_\tau H_\tau^T)\circ(G_\tau G_\tau^T) \quad (29)$$

Each matrix $\Omega_\tau$ is associated only with its own polyharmonic package (its own layer in the cascade) and can be computed independently. Even though the packages may have different numbers of inputs, outputs, and constellation points (and hence different shapes of $Y_\tau$), all matrices $\Omega_1, \Omega_2, \ldots, \Omega_q$ have the same size $r \times r$.

Assume that all matrices (29) have been computed for each package in the cascade. Then from (28) and (29) we obtain

$$B = 2(\Omega_1 + \Omega_2 + \cdots + \Omega_q + \alpha E)^{-1} \Delta L \qquad (30)$$

The vector $B$ found in (30) can then be used independently in each polyharmonic package to compute all elements $\Delta y_{pt}^{(\tau)}$ of the corresponding update matrix $\Delta Y_\tau$ using (25). In this way, the vector $B$ (and the procedure (30) for computing it) plays a synchronizing role among the different packages in the cascade during training, coupling the parameter updates of all packages to one another.

Let us examine (25) more closely. It implies that the element $\Delta y_{pt}^{(\tau)}$ of the matrix $\Delta Y_\tau$ with indices $(p, t)$ is equal to a scalar product involving the p-th column of $H_\tau$, the t-th column of $G_\tau$, and the vector $B$. Consequently, (25) can be written in matrix form as

$$\Delta Y_\tau = \frac{1}{2} H_\tau^T \left( G_\tau \circ (B J_{1,n_\tau}) \right), \qquad (31)$$

where $J_{1,n_\tau}$ is a $1 \times n_\tau$ row vector of ones.

Note that if we drop the factor 2 in (30) and simultaneously drop the factor 1/2 in (31), the final result remains unchanged.

We then obtain the new values of $Y_\tau$ by adding $\Delta Y_\tau$, and update $\Lambda_\tau$ using (11).

Let us now describe the resulting algorithm step by step.

1. **Initialization of the polyharmonic cascade.** Choose the number of packages and the number of functions in each; specify the constellation points (given by the matrices $C_\tau$) and their initial function values (given by the matrices $Y_\tau$). For each package, compute its matrix $U_\tau$ using (9), (4), and (10). Compute the initial values of $\Lambda_\tau$ via (11). Choose the coefficient $\alpha$, which controls the learning rate and will be used in (30). If $\alpha$ is too large, learning may become very slow; if too small, the algorithm may become unstable. (The procedures for initializing $C_\tau$ and $Y_\tau$ are not discussed in this paper and are the subject of a separate study.)
2. **Split the training set into batches and train on them sequentially using the steps below.**
3. **Forward pass for a batch.** Process the batch sequentially through the packages of the cascade, from the first to the last, using (1)-(5). Store the matrices $X_\tau$ and $M_\tau$, and compute and store $H_\tau$ for each package using (13). At the cascade output, compute $\Delta L$, the difference between the desired outputs and the obtained results.

4. **Backward pass (derivatives).** Set $G_q$ to be a column vector of ones. Compute, in reverse, the derivative matrices $G_{q-1}, .., G_1$ using (6)–(8).

5. **Compute the global update direction.** For each package, compute $\Omega_\tau$ using (29). Sum all these matrices and compute $B$ from (30). In (30) a matrix inverse is written explicitly, but in practice it suffices to solve the corresponding linear system for $B$.

6. **Update constellation values and coefficients.** For each package, compute $\Delta Y_\tau$ using the vector $B$ obtained in the previous step and (31). Update $Y_\tau$, and then recompute $\Lambda_\tau$ via (11).

7. **Repeat steps 3–6 for the next batch, and so on.**

The algorithm can be slightly optimized by computing the matrices $\Theta_\tau$ via (6) already in step 3 and storing them instead of $M_\tau$. Then, in step 3, the matrix $K_\tau$ can be computed not via (4), but using

$$K_\tau = M_\tau \circ \left(\Theta_\tau - J_{r,k_\tau}\right) + c \qquad (32)$$

where $J_{r,k_\tau}$ is the $r \times k_\tau$ matrix of ones.

and in step 4 the matrices $\Theta_\tau$ will already be available, so (6) need not be recomputed.

Thus, in steps 1-7 we have outlined the main principles of the training algorithm for the polyharmonic cascade. All expressions are written as operations on two-dimensional matrices, and therefore the algorithm can be implemented efficiently on a GPU.

In contrast to approximate scaling methods for kernel machines, such as the Nyström method (Williams & Seeger, 2001 [15]) or Fast Multipole Methods (Beatson & Greengard, 1997 [4]), the proposed cascade preserves the exact kernel form at every level, without sacrificing global smoothness for computational efficiency.

Formally, the iterative algorithm resembles the Gauss–Newton method (Dennis & Schnabel, 1996 [8]), since at each iteration the nonlinear system is linearized and a least-squares problem is solved. The key difference lies in the choice of optimization space: instead of directly tuning expansion coefficients (which would break the probabilistic interpretation), we optimize the values of hidden functions at fixed constellation points. Likewise, the penalty on changes in these values can be viewed as a generalization of Tikhonov regularization (Tikhonov & Arsenin, 1977 [14]), applied not to weights, but to the intermediate states of the model.

However, as noted at the beginning, in the setting considered so far the cascade has only a single output at the last layer, i.e., the cascade as a whole computes just one function. In practical machine learning tasks, one often needs to compute multiple outputs, for example in classification, where each output corresponds to a separate class.

The algorithm can be adapted to the multi-output setting in several ways.

A straightforward first step is to place a package with multiple outputs at the end of the cascade. Let the number of outputs be $s$. Then $X_q = L$ becomes an $(r \times s)$ matrix. The matrices $G_q$ and $\Delta L$ must then have the same shape (r×s), which no longer fits directly into the single-output algorithm described above.

One possible workaround is to define a scalar target function at the output of the polyharmonic cascade. The values of this function then replace the single cascade output for which the algorithm was originally designed. That is, the values of this scalar function generate the computed and desired vectors $L$ and $L^*$, and hence $\Delta L$ of the required size $(r \times 1)$. The derivatives of this function with respect to $X_q$ fill the entries of the matrix $G_q$ of size $(r \times s)$. Numerical experiments showed that this approach is workable, but it is less efficient and leads to slower training than the methods described below.

Based on numerical experiments, two modifications of the algorithm were developed that allow the proposed training procedure to be adapted to a polyharmonic cascade with multiple outputs.

The first modification of the training algorithm is based on the following rules:

1. For each training example in the batch, we randomly select exactly one output of the cascade and take $l_i$ from that output. The desired value $l^*$ is taken analogously for the same output.
2. The derivative matrices $G_1, G_2, \ldots, G_{q-1}$ are then computed only with respect to the selected outputs. That is, the first row of these matrices contains derivatives with respect to the cascade output chosen for the first training example, the second row contains derivatives with respect to the output chosen for the second training example, and so on.

With this approach, the same training example may (though it need not) appear in the same batch multiple times, if each time a different cascade output is chosen for it.

This can be formalized as follows. We define $G_q$ as a binary matrix

$$G_q \in \{0,1\}^{r \times s}, \text{such that for all } i \in \{1, \ldots, r\} \tag{33}$$

$$\exists! j \in \{1, \ldots, s\}, \text{with } g_{ij}^{(q)} = 1$$

i.e., $G_q$ is a matrix where each row contains exactly one entry equal to 1 and all others are 0. The remaining matrices $G_1, G_2, \ldots, G_{q-1}$ are computed using the same formulas as before.

The matrices $L$ (cascade outputs) and $L^*$ (targets from the training set) now have size $(r \times s)$. From them we obtain $\Delta L$ of size $(r \times 1)$ as

$$\Delta L = \left((L^* - L) \circ G_q\right) J_{s,1} \tag{34}$$

where $J_{s,1}$ is an $s \times 1$ column vector of ones.

The resulting vector $\Delta L$ is then used in the algorithm exactly as before. All subsequent steps of the training procedure remain unchanged.

Let us now consider a second modification of the algorithm that allows us to train a polyharmonic cascade with multiple outputs.

The values computed by the cascade at each of its outputs can be regarded as separate, independent functions, and we can train each of these functions independently. At the same time, we assume that the cascade structure is the same for all outputs: the number of packages, the number of inputs and outputs of each package, and the constellations used in the packages (or the number of points in them) are identical for all functions. In this case, to scale the system to multiple outputs, it suffices to move from operations on two-dimensional matrices to operations on three-dimensional tensors by adding one more axis. This approach is particularly effective for GPU implementations.

Not all matrices in the cascade need to become three-dimensional tensors. The input matrix $X_0$ obviously remains two-dimensional of size $(r \times n_0)$. Likewise, the matrices $C_1, C_2, \ldots, C_q$ can remain two-dimensional (for example, each $C_\tau$ still has size $(k_\tau \times n_{\tau-1})$), and the matrices $U_\tau$ remain $(k_\tau \times k_\tau)$. In principle, one could also turn $C_\tau$ and $U_\tau$ into three-dimensional tensors if, for some reason, one wanted to assign individual constellations per output function in the cascade.

Assume, as in the previous case, that the number of outputs of the cascade is $s$. Then the tensors $\Lambda_\tau$ and $Y_\tau$ have shape $(s \times k_\tau \times n_\tau)$. The tensors $X_\tau$, starting from the output of the first package, have shape $(s \times r \times n_\tau)$.

At the cascade output we obtain a tensor $L = X_q$ of shape $(s \times r \times 1)$, while the target values for the batch, $L^*$, are given as a two-dimensional matrix of size $(r \times s)$. Hence $L^*$ must be transposed and reshaped into a tensor of size $(s \times r \times 1)$, after which we compute the difference to obtain the tensor $\Delta L$. At the same time, $G_q$ is taken to be a tensor of ones of size $(s \times r \times 1)$.

All training computations then proceed according to the same algorithm and the same formulas as in the single-output case, except that $H_\tau, G_\tau, \Omega_\tau$ are now three-dimensional tensors. In (30) one must compute $s$ matrix inverses (or solve $s$ linear systems), yielding a tensor $B$ of shape $(s \times r \times 1)$.

Subsequently, for each package, we compute the tensors $\Delta Y_\tau$ of shape $(s \times k_\tau \times n_\tau)$ using (31), and then update $Y_\tau$ and $\Lambda_\tau$.

In this context, when we speak of operations mixing two-dimensional matrices and three-dimensional tensors – for example, the matrix multiplication of a 2D matrix $U_\tau$ of size $(k_\tau \times k_\tau)$ with a 3D tensor $Y_\tau$ of size $(s \times k_\tau \times n_\tau)$ in (11) – we mean that $s$ independent matrix multiplications are performed between $U_\tau$ and each of the $s$ slices of $Y_\tau$. In practice, machine learning libraries such as PyTorch handle such broadcasting between tensors of different ranks automatically, so when moving from 2D matrices to 3D tensors, the computational code often remains unchanged.

Thus, we have described two modifications of the training algorithm for a polyharmonic cascade with multiple outputs. In numerical experiments, the second variant proved more effective and led to faster and more accurate training, despite the fact that it effectively duplicates the entire cascade and trains it independently for each output (which is still quite efficient on a GPU). The advantage of the first approach is that it can be used when the number of outputs is very large, so that treating each output as a separate independent function would not be practical.

In the numerical experiments, the polyharmonic cascade as a machine learning architecture – including the training algorithm described in this paper – was implemented in Python using the PyTorch library.

It was confirmed that the proposed algorithm is suitable for training a polyharmonic cascade consisting both of only a few layers and of dozens of successive polyharmonic packages (from a few hundred up to tens of millions of trainable parameters).

Below we describe an example of training a polyharmonic cascade on the MNIST dataset [11] (loaded via the torchvision library).

When training on MNIST, no account was taken of the underlying 28×28 image structure (as is done in convolutional networks). Instead, the inputs were simply vectors of length 784 (i.e., 784 cascade inputs).

For this dataset we used a polyharmonic cascade consisting of four successive packages. The first package took 784 inputs and computed 100 functions. The second package took 100 inputs and computed 20 functions. The third package took 20 inputs and computed 20 functions. The fourth package took 20 inputs and computed a single function. Using the second modification method described above, this entire structure was then scaled to 10 outputs.

Taking into account all constellations, the total number of trainable parameters in the cascade (i.e., all entries of the matrices $Y_\tau$) was about 1.6 million.

Training was carried out on an NVIDIA GeForce RTX 3070 GPU.

The training curve is shown in Figure 2.

From the curve one can see that the polyharmonic cascade effectively learned almost immediately in the first epoch, reaching a test accuracy of 97.52%. After 2–3 epochs the test accuracy stabilized. This indicates the absence of any pronounced overfitting as training proceeds (at least for this task).

After 10 epochs the accuracy reached 98.3%.

When repeating training with different random initializations, the final accuracy varied slightly, typically in the range 98.2–98.4%, but the qualitative shape of the learning curve remained almost identical.

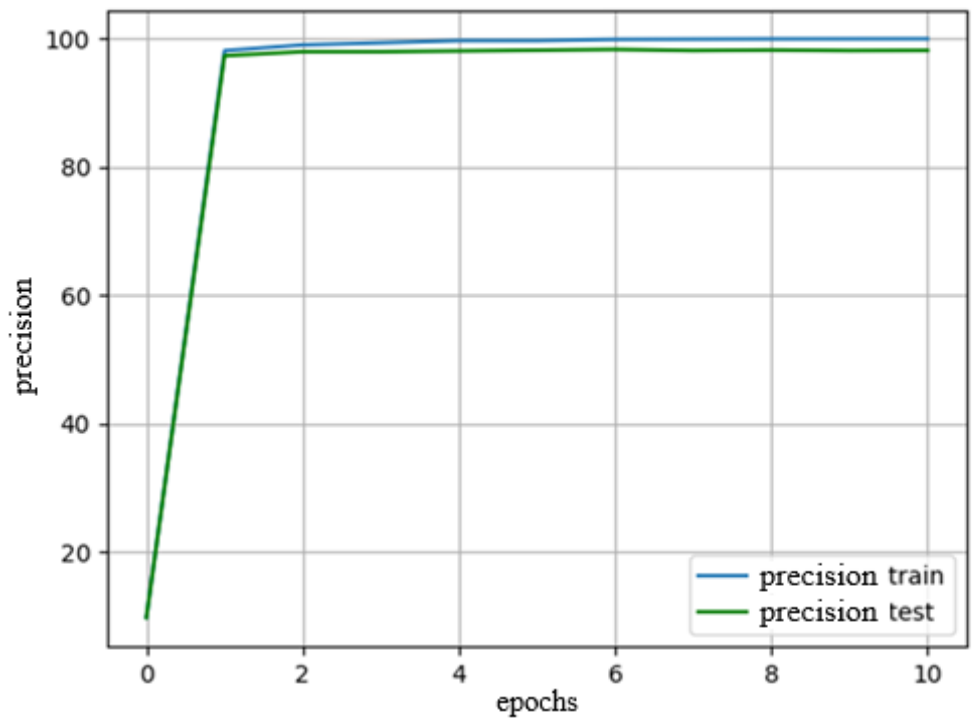

Figure 2. Training curve on the MNIST dataset

The training time per epoch was about 1.5 seconds.

**Conclusion.**

The proposed training method is not merely a computational trick but a logical continuation of the probabilistic paradigm introduced in the first paper of the series [1]. The refusal to apply gradient descent directly to the coefficients $\lambda$ is motivated by their status as Lagrange multipliers rather than free parameters. Switching to optimization over function values at constellation points preserves the theoretical integrity of the model even in a deep composition. This makes the polyharmonic cascade an architecture in which each layer and the training procedure itself have a probabilistic justification derived from first principles.

**Acknowledgments.**

The author thanks the anonymous reviewers for valuable feedback on earlier versions of this work.

**References**

1. Bakhvalov, Y. N. (2025). Solving a Machine Learning Regression Problem Based on the Theory of Random Functions. arXiv. https://doi.org/10.48550/arXiv.2512.12731
2. Bakhvalov, Y. N. (2025). Polyharmonic Spline Packages: Composition, Efficient Procedures for Computation and Differentiation. arXiv. https://doi.org/10.48550/arXiv.2512.16718
3. Bartsev, S. I., & Okhonin, V. A. (1986). Adaptivnye seti obrabotki informatsii [Adaptive information-processing networks]. Krasnoyarsk: Institute of Physics, Siberian Branch of the USSR Academy of Sciences. Preprint No. 59B, 20 pp. [in Russian]
4. Beatson, R. K., & Greengard, L. (1997). A short course on fast multipole methods. In M. Ainsworth, J. Levesley, W. Light, & M. Marletta (Eds.), *Wavelets, Multilevel Methods and Elliptic PDEs* (pp. 1–37). Oxford University Press.
5. Bookstein, F. L. (1989). Principal warps: Thin-plate splines and the decomposition of deformations. *IEEE Transactions on Pattern Analysis and Machine Intelligence, 11*(6), 567–585. https://doi.org/10.1109/34.24792
6. Buhmann, M. D. (2003). *Radial Basis Functions: Theory and Implementations.* Cambridge University Press.
7. Damianou, A., & Lawrence, N. (2013). Deep Gaussian processes. In *Proceedings of the Sixteenth International Conference on Artificial Intelligence and Statistics (AISTATS)* (pp. 207–215).
8. Dennis, J. E., & Schnabel, R. B. (1996). *Numerical Methods for Unconstrained Optimization and Nonlinear Equations.* SIAM.
9. Duchon, J. (1977). Splines minimizing rotation-invariant semi-norms in Sobolev spaces. In W. Schempp & K. Zeller (Eds.), *Constructive Theory of Functions of Several Variables* (pp. 85–100). Springer. https://doi.org/10.1007/BFb0086566

10. Harder, R. L., & Desmarais, R. N. (1972). Interpolation using surface splines. *Journal of Aircraft, 9*(2), 189–191.

11. LeCun, Y., Bottou, L., Bengio, Y., & Haffner, P. (1998). Gradient-based learning applied to document recognition. *Proceedings of the IEEE, 86*(11), 2278–2324. https://doi.org/10.1109/5.726791

12. Pugachev, V. S. (1960). Teoriya sluchainykh funktsii i ee primenenie k zadacham avtomaticheskogo upravleniya [Theory of Random Functions and its Application to Problems of Automatic Control] (2nd ed., rev. and enl.). Moscow: Fizmatlit. [in Russian]

13. Rumelhart, D. E., Hinton, G. E., & Williams, R. J. (1986). Learning internal representations by error propagation. In D. E. Rumelhart & J. L. McClelland (Eds.), *Parallel Distributed Processing: Explorations in the Microstructure of Cognition, Vol. 1* (pp. 318–362). MIT Press.

14. Tikhonov, A. N., & Arsenin, V. Y. (1986). Metody resheniya nekorrektnykh zadach [Methods for Solving Ill-Posed Problems] (3rd ed., rev.). Moscow: Nauka. [in Russian]

15. Williams, C. K. I., & Seeger, M. (2001). Using the Nyström method to speed up kernel machines. In T. K. Leen, T. G. Dietterich, & V. Tresp (Eds.), *Advances in Neural Information Processing Systems 13 (NIPS)* (pp. 682–688). MIT Press.

16. Wilson, A. G., Hu, Z., Salakhutdinov, R., & Xing, E. P. (2016). Deep kernel learning. In *Proceedings of the 19th International Conference on Artificial Intelligence and Statistics (AISTATS)* (pp. 370–378).